%% file: main.tex
\documentclass[
]{ceurart}

\usepackage{listings}
\usepackage{url}
\usepackage{tikz}
\usepackage{pgfplots}
\usepackage{pgfplotstable}
\usepackage{graphicx}
\usepackage{subcaption}
\pgfplotsset{compat=1.18}
\usepackage{placeins}
\usetikzlibrary{arrows.meta, positioning}

\begin{document}

\copyrightyear{2026}
\copyrightclause{Copyright for this paper by its authors.
  Use permitted under Creative Commons License Attribution 4.0
  International (CC BY 4.0).}

\conference{SeMatS 2026: The 3rd International Workshop on Semantic Materials Science co-located with the 25th International Semantic Web Conference (ISWC 2026), October 25th, 2026, Bari, Italy}

\title{Automatically Building and Updating a Knowledge Graph of MLIP Models}

%% \tnotemark[1]
%% \tnotetext[1]{You can use this document as the template for preparing your
%%  publication. We recommend using the latest version of the ceurart style.}

\author[1]{Alexis Beer}[%
%orcid=0000-0002-0877-7063,
email=alexis.beer@loria.fr,
%url=https://yamadharma.github.io/,
]
\fnmark[1]
\address[1]{MosAIk Team, LORIA, Université de Lorraine, CNRS, Nancy, France}
\address[2]{CRET AIREL, Université de Lorraine, Nancy, France}

\author[1,2]{Liudmyla Klochko}[%
orcid=0000-0003-2468-8738,
email=liudmyla.klochko@loria.fr,
%url=https://kmitd.github.io/ilaria/,
]
\fnmark[1]
\cormark[1]
%\address[3]{Vrije Universiteit Amsterdam, De Boelelaan 1105, 1081 HV Amsterdam, The Netherlands}

\author[1,2]{Mathieu d'Aquin}[%
orcid=0000-0001-7276-4702,
email=mathieu.daquin@loria.fr,
% url=http://conceptbase.sourceforge.net/mjf/,
]
\fnmark[1]

\cortext[1]{Corresponding author.}
\fntext[1]{These authors contributed equally.}

\begin{abstract}
Complementing the many efforts in providing semantic representations of concepts, notions, and entities in materials science, we report and illustrate a process by which we can automatically build a knowledge graph of the fast evolving field of machine learning applied to the prediction of material properties, focusing on MLIP (Machine Learning Interatomic Potential). This LLM-based process relies on multiple steps, from information extraction in documents and articles to a validation loop using SHACL constraints to detect and correct errors. It is carried out on a model-by-model basis, focusing on the consistency of representation, therefore enabling an iterative construction where the addition of new models is facilitated. We illustrate the process by showing a few interesting aspects that can be queried from a knowledge graph built from the models listed in the Matbench Discovery leaderboard. 
\end{abstract}

\maketitle

\section{Introduction}

The application of AI methods, especially deep learning-based models, to materials science is progressing very quickly. In just the last couple of years, we have seen such applications evolve from simple prediction models of specific physical properties to the use of Physics-Informed Neural Networks (PINNS)~\cite{Raissi2019} and the development of specific types of GNNs. One of the fastest areas of development, at the time of writing, concerns foundational machine learning interatomic potential (MLIP)~\cite{Jacobs2025} or force field (MLFF) models that are trained on millions of numerical simulations, mostly based on atomic structures, and can be used/adapted/fine-tuned to enable many applications. This can be seen particularly through the Matbench Discovery leaderboard\footnote{\url{https://matbench-discovery.materialsproject.org/}}, where the top ranked models are GNN-based MLIP models, with new competitive models being introduced on a regular basis. 

Those models have the potential to fundamentally change the way materials science operates. However, as the field evolves so quickly and new models appear constantly, it also becomes harder to keep track of those developments and exploit them effectively. What new models have been proposed? What distinguishes them from already existing ones? How have they been trained? Are they suitable for the specific materials/conditions relevant to a particular application? Answering these questions requires continuous monitoring of benchmark results, leaderboards, and the rapidly growing scientific literature.

In this paper, we propose summarizing all the relevant information about those models into a constantly evolving knowledge graph to effectively support this monitoring. Such a process requires guidance from an ontology that is tailored for the representation of deep learning-based models suitable for describing MLIP models, their training, and evaluation. Our goal is to obtain a detailed description of those aspects of MLIP models, with information that is currently scattered across unstructured sources. Indeed, considering the rapid pace of development of such models, we cannot assume that a manual population of the knowledge graph would be sufficient. For this reason, we propose a Large Language Model (LLM)-based information extraction and knowledge graph construction process. This process is guided by a specifically created ontology to obtain information that instantiates it with data obtained from articles and other textual resources\footnote{in MatBench Discovery, models are described through pdf articles, a YAML file that provides some basic information, and which also links to datasets relevant to the evaluation of models. This is what we mean by articles and textual resources.} that describe each model. One particularity of the process we propose is that we not only include validation of the resulting knowledge graph through a suite of tests implemented by SHACL (Shapes Constraint Language) constraints~\cite{Pareti2019-rh}, but we also include an LLM-based approach to correct errors detected by this process.

This paper is structured as follows: first, we summarize the elements of related works that have been useful in the development of our knowledge graph construction process. Then, we provide an overview of the approach and a section for each of the main components of the process: the necessary ontology to be used by the process, the LLM-based information extraction method, and the LLM- and SHACL-based error correction method. Finally, we conclude on the state of the created knowledge graph for the models listed under the Matbench Discovery leaderboard. All the code, ontologies and results for this process are available in a public repository\footnote{\url{https://github.com/alexisbe-github/Matbench-onto}}. 

\section{Related work}

%% Ontologies for materials and KGs for materials

 Materials science is a highly interdisciplinary field that combines aspects of materials at different scales and from different points of view: theoretical, experimental, application, etc. As a result, many works have focused on providing ways to semantically describe and structure various aspects of the field, including studies that relied on applying ontologies and knowledge graphs to materials science~\cite{Venugopal2024,Zhang2024}.  The goal of this section is not to provide a complete account of those works. Generally, they tend to be motivated by similar arguments to those that motivate the construction of ontologies and knowledge graphs in other sciences~\cite{ding2026bridging}. In particular, knowledge graphs and ontologies support semantic interoperability, and materials science, like many other sciences, is based on information that varies in form and nature, including measurements, experiments, and computations described in databases, articles, and other repositories. Many applications of knowledge graphs and ontologies in the field have therefore been related to the integration of heterogeneous data sources~\cite{zhang2015survey,norouzi2024landscape}. Closer to the work in this paper, a number of studies have started looking into automating the extraction of information about materials (and related concepts) from articles and other textual resources that could be used to build knowledge graphs and ontologies. These include works to create embeddings of materials from the literature, such as the mat2vec model~\cite{tshitoyan2019unsupervised}. They also include specific NLP pipelines adapted to the extraction of entities related to materials~\cite{weston2019named}. More recently, automatic processes for extracting entire knowledge graphs~\cite{zhang2024materials} have also started appearing, including those relying on LLMs~\cite{ye2024construction,li2026extracting}. In this work, we focus on building a knowledge graph of machine learning models for materials, especially focusing on MLIP models, which, to the best of our knowledge, has not yet been explored. 

%% Ontologies and KGs for models

Since our focus is on machine learning models for materials science, related works also include ontologies and knowledge graphs that are dedicated to the representation of machine learning models. Surprisingly, relatively few works have been dedicated to this topic. The ML Schema~\cite{publio2018ml} is one example, promoted by a dedicated W3C community group, that provides a set of classes and properties for describing machine learning experiments. It relies on other ontologies to describe models, their implementations, and evaluations at a high level. Similarly, SML (Semantic Machine Learning Model Ontology~\cite{kallab2023sml}) describes machine learning models, the datasets on which they rely, and notions of evaluation. It is more focused on usability for describing practical applications rather than experiments. Finally, FAIR4ML\footnote{\url{https://github.com/RDA-FAIR4ML/FAIR4ML-schema}} is an ontology-based extension of Schema.org specifically targeting the FAIRness of ML model documentation, with classes and properties for evaluation metrics, intended applications, and tasks. There are a few other ontologies, vocabularies, and thesauri that we could have cited here. As described later, our objective is to describe the fast evolving MLIP models used to predict the physical properties of materials in a granular way, so that those descriptions could be used, for example, for model recommendation. This implies representing their architectures, training parameters, versions, variants, evaluations, and specific prediction results. Since we focus here on building a proof-of-concept process for information extraction that is guided by an ontology, we rely in this first step on a specifically made ontology and do not describe the methodology for creating it in details. We aim to refine the ontology itself, and align it with existing ones in materials science and on machine learning models, as part of our future work. 

\section{Overview of the approach}
\label{sec:overview}

% ============================================================
%  COLOR PALETTE -- edit these three lines to re-theme the
%  whole diagram (pipeline / process / LLM+graph elements)
% ============================================================
\definecolor{colStructure}{RGB}{0,0,0}      % ontology boxes, LLM boxes, plain arrows
\definecolor{colProcess}{RGB}{224,102,25}   % orange process/extraction boxes
\definecolor{colGraph}{RGB}{37,72,212}      % LLM(graph) box, KG icons, blue arrows

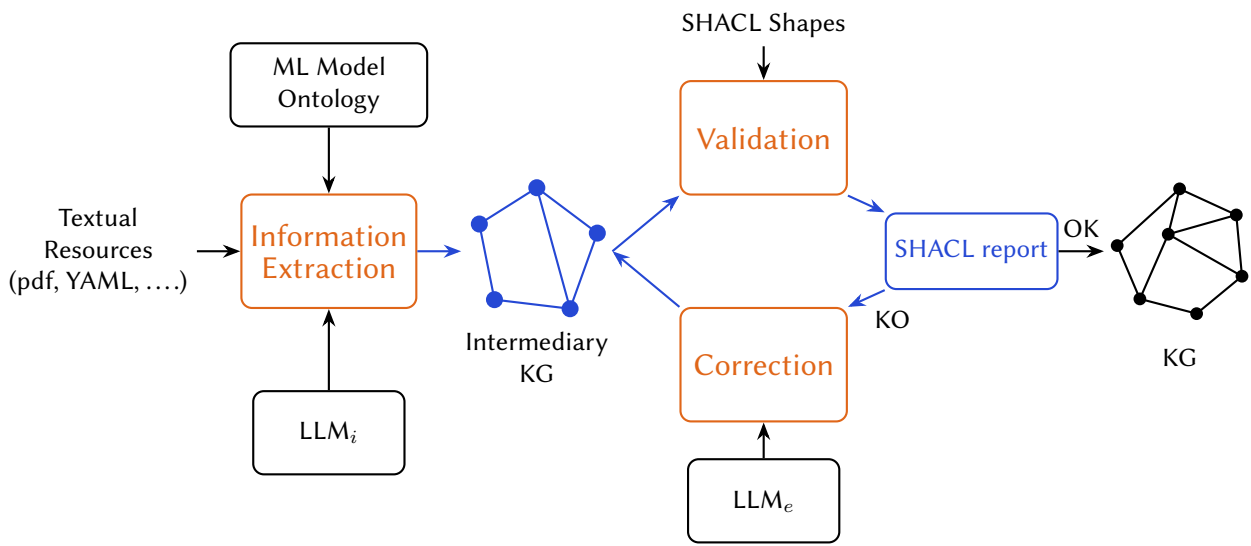
\begin{figure}
    \centering

\begin{tikzpicture}[
  >=Stealth,
  every node/.style={font=\small},
  blackbox/.style={rectangle, rounded corners, draw=colStructure, thick,
                    minimum width=2.6cm, minimum height=1.1cm, align=center},
  orangebox/.style={rectangle, rounded corners, draw=colProcess, thick,
                     text=colProcess, minimum width=2.2cm, minimum height=1.5cm,
                     align=center},
  bluebox/.style={rectangle, rounded corners, draw=colGraph, thick, text=colGraph,
                   minimum width=1.3cm, minimum height=1cm, align=center},
  placeholder/.style={text=colStructure, align=center},
  label/.style={align=center},
]

%======================================================
% LEFT BRANCH: text extraction into an "instance" KG
%======================================================
% \node[placeholder]                  (toplabel) at (0,5.6)   {[corpus type] $\rightarrow$ KG};
\node[blackbox]                     (ontology) at (-1.25,4)      {ML Model\\Ontology};
\node[orangebox]                    (extract)  at (-1.25,1.8)    {\large Information\\ \large Extraction};
\node[label, align=center]          (textres)  at (-4.3,1.8) {Textual\\Resources\\(pdf, YAML, \dots.)};
\node[blackbox, minimum width=2cm]  (llmi)     at (-1.25,-0.6)   {LLM$_i$};

\coordinate (ikg) at (2.5,1.8);
\coordinate (iikg) at (0.5,1.8);
\node[placeholder] at (1.5,0.4) {Intermediary\\KG};

%======================================================
% RIGHT BRANCH: schema/shape-guided validation into final KG
%======================================================
\node[placeholder]                  (shacl)    at (4.5,4.8) {SHACL Shapes};
\node[orangebox]                    (vocab)    at (4.5,3.3)   {\large Validation};
\node[bluebox]                      (llmv)     at (7.25,1.8) {SHACL report};
\node[placeholder]                  (obslabel) at (8.7,2.1) {OK};

\coordinate (kg) at (9,1.8);
\node[label] at (10,0.4) {KG};

\node[orangebox]                    (corr)     at (4.5,0.3) {\large Correction};
\node[placeholder]                  (ontoname) at (6.2,0.9) {KO};
\node[blackbox, minimum width=2cm]  (llme)     at (4.5,-1.5)  {LLM$_e$};

%======================================================
% ARROWS -- left branch
%======================================================
%\draw[->, thick] (toplabel) -- (ontology.north);
\draw[->, thick] (ontology.south) -- (extract.north);
\draw[->, thick] (textres.east) -- (extract.west);
\draw[->, thick] (llmi.north) -- (extract.south);
\draw[->, thick, colGraph] (extract.east) -- (iikg);

%======================================================
% ARROWS -- right branch
%======================================================
\draw[->, thick] (shacl) -- (vocab.north);
\draw[->, thick, colGraph] (vocab.south east) -- (llmv.north west);
\draw[->, thick, colGraph] (llmv.south west) -- (corr.north east);
\draw[->, thick, colGraph] (ikg) -- (vocab.south west);
\draw[->, thick, colGraph] (corr.north west) -- (ikg);
\draw[->, thick] (llmv.east)  -- (kg);
\draw[->, thick] (llme.north) -- (corr.south);

%======================================================
% small "graph" glyphs -- each is a little node-link graph
% rather than a plain diamond outline
%======================================================
% iKG: small/instance graph (5 nodes, few links)
\begin{scope}[shift={(1.5,1.8)}, scale=2.0, colGraph, thick]
  \coordinate (a) at (0,0.42);
  \coordinate (b) at (0.4,0.12);
  \coordinate (c) at (0.22,-0.38);
  \coordinate (d) at (-0.28,-0.32);
  \coordinate (e) at (-0.38,0.18);
  \draw (a) -- (b);
  \draw (b) -- (c);
  \draw (c) -- (d);
  \draw (d) -- (e);
  \draw (e) -- (a);
  \draw (a) -- (c);
  \foreach \p in {a,b,c,d,e}{\fill (\p) circle (1.6pt);}
\end{scope}

% KG: larger/final graph (7 nodes, denser links)
\begin{scope}[shift={(10,1.8)}, scale=1.5, thick]
  \coordinate (a) at (0,0.55);
  \coordinate (b) at (0.5,0.32);
  \coordinate (c) at (0.55,-0.22);
  \coordinate (d) at (0.15,-0.55);
  \coordinate (e) at (-0.35,-0.42);
  \coordinate (f) at (-0.55,0.05);
  \coordinate (g) at (-0.1,0.15);
  \draw (a) -- (b);
  \draw (b) -- (c);
  \draw (c) -- (d);
  \draw (d) -- (e);
  \draw (e) -- (f);
  \draw (f) -- (a);
  \draw (g) -- (a);
  \draw (g) -- (c);
  \draw (g) -- (e);
  \draw (g) -- (b);
  \foreach \p in {a,b,c,d,e,f,g}{\fill (\p) circle (1.6pt);}
\end{scope}

\end{tikzpicture}

    \caption{Overview of the automated knowledge graph construction process.}
    \label{fig:overview}
\end{figure}

Figure~\ref{fig:overview} provides an overview of the process presented in this paper. The process starts with an information extraction step that takes as input textual resources describing a particular model. Here, pdf and YAML are mentioned, as those are the main formats for describing the models participating in the leaderboard on the Matbench Discovery portal. Information extraction here is carried out through the use of a LLM (see Section~\ref{sec:extraction}) and is guided by our ML Model Ontology (see Section~\ref{sec:ontology}) in the sense that it is aimed not only at producing a knowledge graph that is structured according to the ontology, but it also relies on the ontology as a resource to identify information to extract (ask the right questions). A key innovation in our process is the validation-correction loop (see Section~\ref{sec:correction}). This process takes SHACL Shapes~\cite{Pareti2019-rh} that act as integrity constraints or unit tests for the created knowledge graph of a given model. The knowledge graph is therefore validated against those constraints. If any inconsistency is detected, another LLM is used, taking as input the original resources, the created knowledge graph, and the SHACL validation report, and is instructed to find ways to rectify the inconsistency. This process is carried out in a loop until all inconsistencies have been removed, ending the process with a cleanly structured knowledge graph for the considered model. That the knowledge graph for a model is defined according to our ontology and validates our SHACL constraints ensures that it can be integrated into a global knowledge graph of all models. In this way, the global knowledge graph can be easily updated when a new model appears. However, the absence of SHACL violations only demonstrates conformance with the explicitly defined constraints and does not guaranty the full completeness or factual correctness of the resulting knowledge graph.

\section{An Ontology of ML Models}
\label{sec:ontology}

As mentioned before, a fine-grained ontology is required for the description of machine learning models to be used for the knowledge graph of MLIP models. Creating a formally validated, aligned, and reusable ontology for describing MLIP models is an entire work in itself, which we choose to leave for future work. The core aspects of the methodology to build it include the requirements for it to describe models, including all the ways in which they are present on Matbench Discovery, detailing their architecture, the data on which they were trained, and their evaluation. A key consideration was also to take into account that MLIP models often evolve and may be grouped into model families, variants, etc. Finally, while we include evaluation results, with metrics, tasks, and benchmarks to be usable for model recommendations, the specific predictions and expected values on individual test inputs must also be included. 

Following these requirements, the ontology is modularized into three main components: the core that defines models, their relations, and architectures; the training module that describes training strategies, hyperparameters, and data; and the evaluation module, which describes both global metrics run on benchmarks and results from individual inputs.

%% core 
\subsection{The core ML Model Ontology}

\begin{figure}
    \centering
    \includegraphics[width=0.75\linewidth]{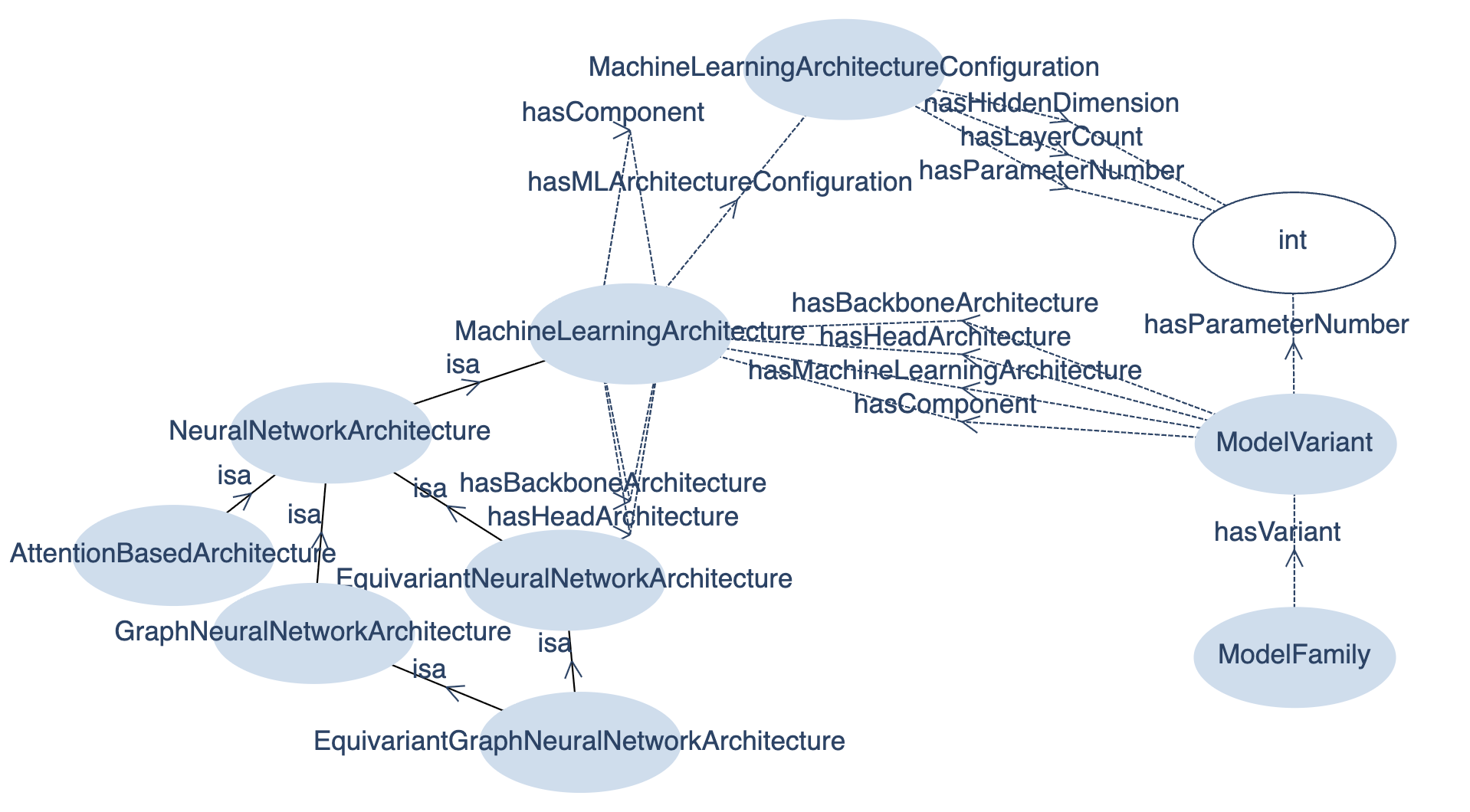}
    \caption{Diagram of an extract from the core (architecture) ML Model Ontology. (\texttt{isa} means \texttt{rdfs:subClassOf})}
    \label{fig:coreonto}
\end{figure}

Figure~\ref{fig:coreonto} visually summarizes the core ML model ontology. As can be seen, the main class is called ModelVariant because it is rare for what we would usually refer to as an MLIP model to exist in only one version. It is however important that when we describe a model, we identify the exact version we are describing. For the more general notion of a model, the one for which there can be multiple variants, we use the class ModelFamily. As can be seen, besides this aspect, the core ontology focuses mostly on describing models (variants) through their architecture. The architecture class itself enables the description of architectures recursively, through components, and specific properties are defined for components that take a particular role (backbone, head, etc.) In this way, we can, for example, describe a GNN model with an MLP-based classification head. The architecture can also be associated with a configuration that enables specifying basic descriptors for the architecture (or its components), such as the number of parameters or the number of layers. 

%% training
\subsection{The ML Training Module}

\begin{figure}
    \centering
    \includegraphics[width=0.75\linewidth]{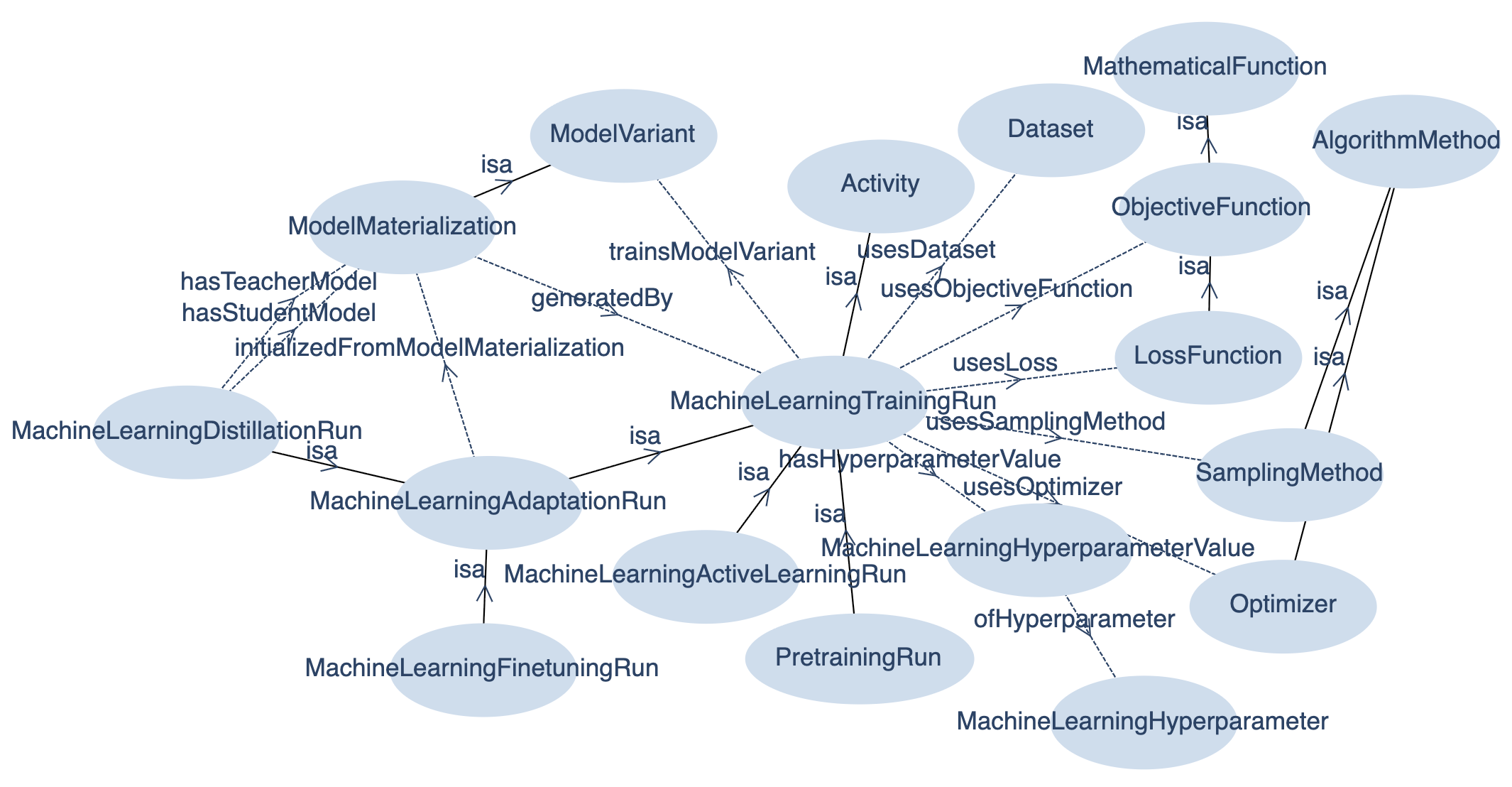}
    \caption{Diagram of an extract from the training module. (\texttt{isa} means \texttt{rdfs:subClassOf})}
    \label{fig:trainingonto}
\end{figure}

The training module, as summarized in Figure~\ref{fig:trainingonto}, mostly focuses on describing the process by which a model variant has been produced. The central class is therefore MachineLearningTrainingRun, which connects to the ModelVariant class of the core ontology. It also includes a number of classes and properties to describe the setup of the training run, with aspects such as the dataset, the loss function, the optimizers, and the hyperparameters used. A large part of the module is also dedicated to describing different training strategies, some of which, such as distillation or fine-tuning, take existing models (materializations, which are concrete representations/implementations of model variants) as input.

%% evaluation
\subsection{The ML Evaluation Module}

\begin{figure}
    \centering
    \includegraphics[width=0.75\linewidth]{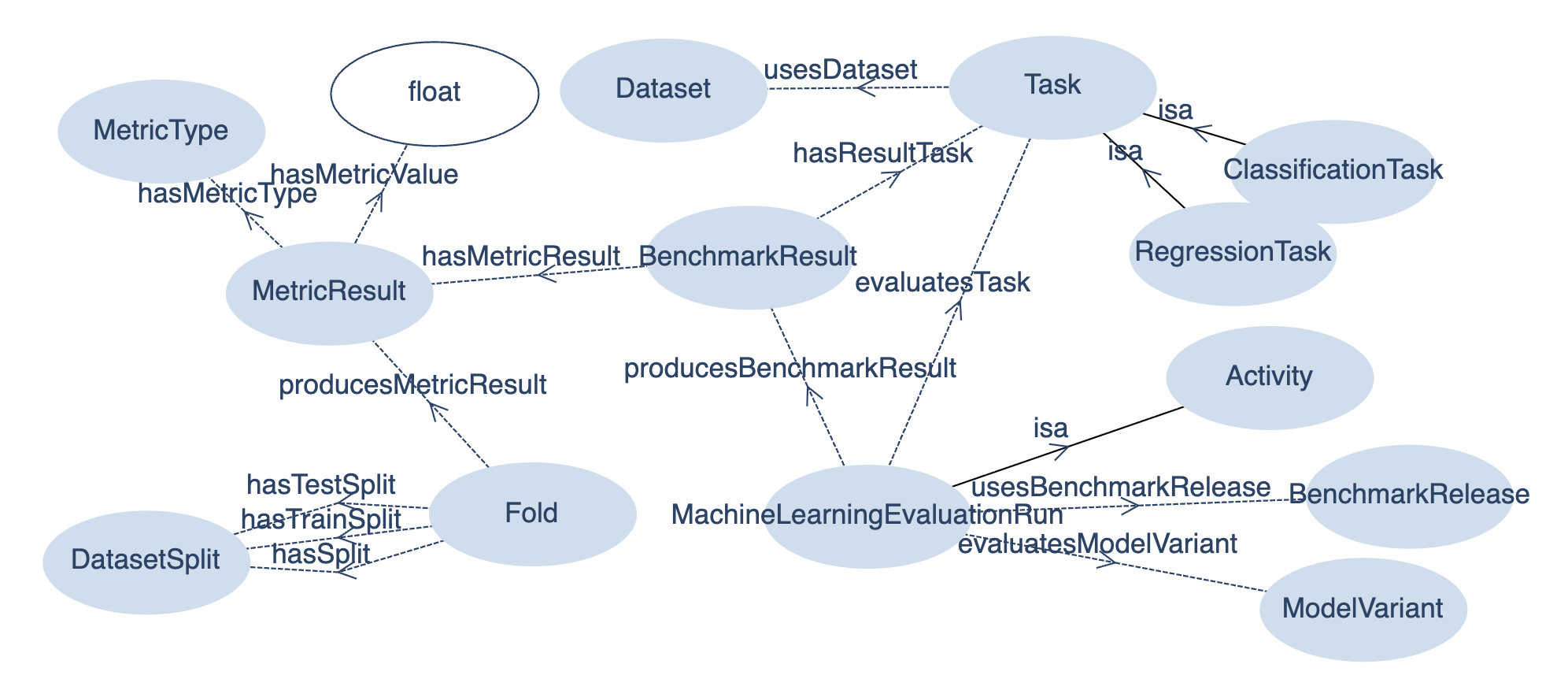}
    \caption{Diagram of an extract from the evaluation module. (\texttt{isa} means \texttt{rdfs:subClassOf})}
    \label{fig:evalonto}
\end{figure}

The evaluation module is summarized in Figure~\ref{fig:evalonto} and includes classes for the tasks being carried out and the metrics used to measure the performance of models. The central class is the MachineLearningEvaluationRun, which represents the evaluation of a ModelVariant on a particular task. Each evaluation run produces a BenchmarkResult, which groups the resulting MetricResult instances.

One aspect that is not represented in the figure is the evaluation of results on individual inputs. This aspect is relatively simple, as it involves attaching a prediction result to a model variant that provides an identification of the input, as well as the predicted value and the expected value. Based on this, evaluation metrics on subsets of the datasets can be easily re-calculated. 

\section{LLM-based Process for Extracting Model Details from Documents}
\label{sec:extraction}

%%% JSON, then translation

As already described in Section~\ref{sec:overview}, the information extraction process is designed to extract the part of the knowledge graph around one particular model (or rather, one model variant). The starting point is therefore textual documents that are supposed to collectively contain all the details needed to fully describe a model according to our ontology (including its training and evaluation). In the rest of the paper, the process will be illustrated using resources available through the Matbench Discovery portal. For each model, a pdf file serves as the reference article in which the authors have described the model, and a YAML file contains the Matbench description of the model, including the results of evaluations. The results of individual predictions are treated separately, as they can be interpreted through a simple script without requiring the use of a LLM. 

In order to guide the LLM in identifying the relevant pieces of information, a JSON file is created that maps each property of relevant objects around the model variants to a question in English for the LLM to be queried with. In order to facilitate the process, taking into consideration that the syntax of RDF (Resource Description Framework\footnote{\url{https://www.w3.org/RDF/}}), which we ultimately want to use for our knowledge graph, is not as familiar as that of JSON, the process is divided into two steps: 

\begin{enumerate}
    \item \textbf{Information extraction.}
    The LLM is first asked to extract all the information corresponding to all those questions to produce one JSON file reflecting this structure and including the extracted information values.
    \item \textbf{Knowledge representation.}
    In a separate process, the extracted JSON file is converted into RDF
    using the ontology as a semantic reference. The LLM is provided with
    both the JSON content and the relevant ontology classes and properties and is asked to generate RDF statements that conform to the ontology.
\end{enumerate}

%%% Interesting point: evidence extraction and use of RDF-Star.

One interesting aspect of the process described here is that it not only obtains the extracted information (the values of properties directly or indirectly attached to the model variant individual in the knowledge graph). The LLM is also prompted to provide evidence for every extracted piece of information. Such evidence takes the form of a reference to the paragraph or section of the document in which the information can be found, or a quote from the document. For example, for the equiformer-v3-mp model\footnote{\url{https://huggingface.co/mirror-physics/equiformer_v3}}, the initial JSON-based extraction might contain the following information regarding the number of layers:
\begin{verbatim}
" architecture": {
   ...
   "layer_count": {
        "value": 7,
        "evidence": "PDF Appendix Table 10: Number of Transformer blocks = 7 
                     for MPtrj training"
    }, 
    ...
\end{verbatim}
clearly indicating that the architecture of the model contains 7 layers, representing 7 blocks of transformers, and that this information is available in Table~10 of the appendix of the provided pdf document. 

Those pieces of evidence represent crucial information regarding the provenance of various elements of the knowledge graph, which can be used to verify and validate the values obtained for various properties and to identify their sources. We therefore integrate it directly into the RDF representation of the information through the use of RDF-star\footnote{\url{https://w3c-cg.github.io/rdf-star/cg-spec/editors_draft.html}}. Indeed, RDF-star enables the integration of triples about triples in RDF. In this case, the information about the number of layers of the model will be represented as a triple between its architecture's configuration and the number 7. The source of this information can therefore be added as an annotation of this triple in the following way in RDF-star: 
\begin{verbatim}
<< mlip:equiformer_v3_mp_equiformerv3_architecture_configuration 
                   mlo:hasLayerCount "7"^^xsd:integer >> 
                            prov:wasDerivedFrom "PDF Appendix Table 10: 
                            Number of Transformer blocks = 7 for MPtrj training" ; 
                            dcterms:source <https://arxiv.org/pdf/2604.09130.pdf> .
\end{verbatim}
Hence, the produced knowledge graph can be queried not only for detailed information about each model but also for the precise origin or provenance of the information. 

\section{SHACL-based Validation and Correction of the Extracted Knowledge Graph}
\label{sec:correction}

%%%

We cannot naturally assume, even with large, recent LLMs, that the information extracted will always be correct and adequately represented. Including the source of each piece of information in every triple, as described above, can help with spot-checking a subset of the generated triples for accuracy. However, carrying out such verifications for every aspect of the knowledge graph would be counterproductive to our objective of automating the construction of the knowledge graph. For this reason, we also include automatic validation of the produced knowledge graph for each model, which enables us to set up another LLM-based process for error correction. 

For this process, we naturally rely on SHACL. Indeed, SHACL enables writing constraints (shapes) that should be verified by the knowledge graph and ensures that they indeed are. Based on our design of the ontology as well as the recommendations from SoCK~\cite{luthfi2022sock}, we built a number of SHACL shapes to act as unit tests, completeness tests, and integrity constraints for the constructed knowledge graph. For a given model, the created knowledge graph will therefore be validated, leading to a SHACL report listing all the violated constraints. Using this report, which indicates errors in the issue of representation and the original sources, a second LLM (which does not have to be the same as the first) is therefore instructed to correct the identified issues. This process is repeated in a loop until no new issues are identified by SHACL, leading to a knowledge graph that is valid according to the established constraints. For example, in the initially generated knowledge graph for AlchemBERT\footnote{\url{https://gitee.com/liuxiaotong15/alchemBERT}}, the individual \texttt{alchembert\_pretraining} did not specify the dataset used during pre-training. This violated a \texttt{sh:minCount 1} constraint requiring every \texttt{train:PretrainingRun} to have at least one \texttt{train:usesDataset} relation. Based on the SHACL report and the original sources, the repair LLM added the following missing triple:
\begin{verbatim}
    trind:alchembert_pretraining train:usesDataset dataind:mp_2022 .
\end{verbatim}
The subsequent SHACL validation confirmed that this violation had been resolved.

\section{Results and Discussion}
\label{sec:discussion}

%% some stats of the knowledge graph

We applied the process described above to all the models of the Matbench Discovery portal using Meituan's LongCat-2.0-Preview LLM\footnote{\url{https://www.longcatai.org/models/longcat-2}} for both information extraction ($LLM_i$ in Figure~\ref{fig:overview}) and error correction ($LLM_e$ in Figure~\ref{fig:overview}).\footnote{The reason for using this LLM was that it was free in stealth mode on OpenRouter under the name OWL Alpha at the time and achieved very high performance.} This resulted in a knowledge graph containing 49 model variants from 31 model families. 

%% perf vs number of parameters 
\begin{figure*}[!ht]
    \centering
    \begin{subfigure}{0.95\textwidth}
        \centering
        \includegraphics[width=\linewidth]
        {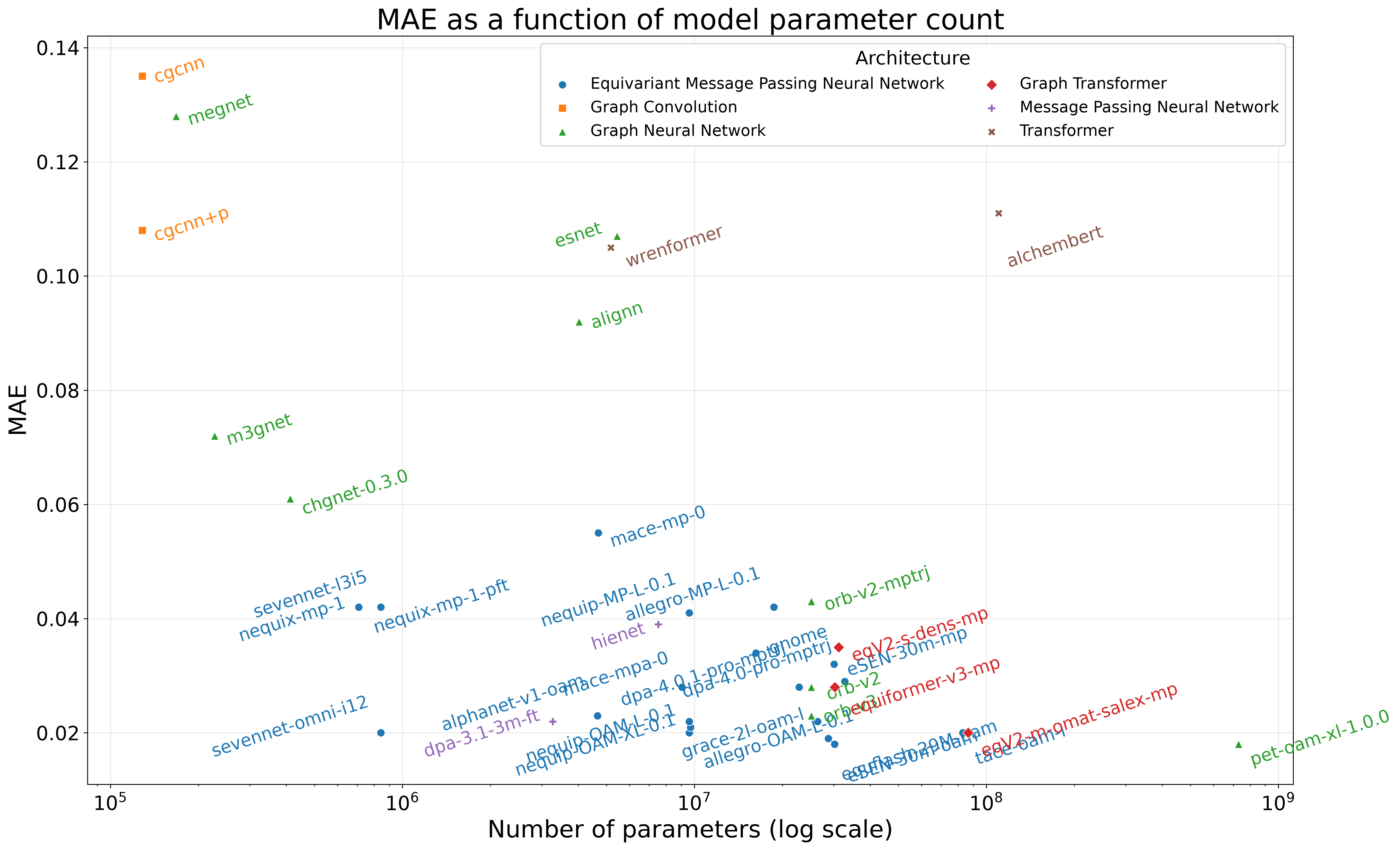}
        \caption{Model performance as a function of the number of parameters.
        Colors indicate the architecture type.}
        \label{fig:performance-parameters}
    \end{subfigure}

    \vspace{0.5cm}

    \begin{subfigure}{0.95\textwidth}
        \centering
        \includegraphics[width=\linewidth]
        {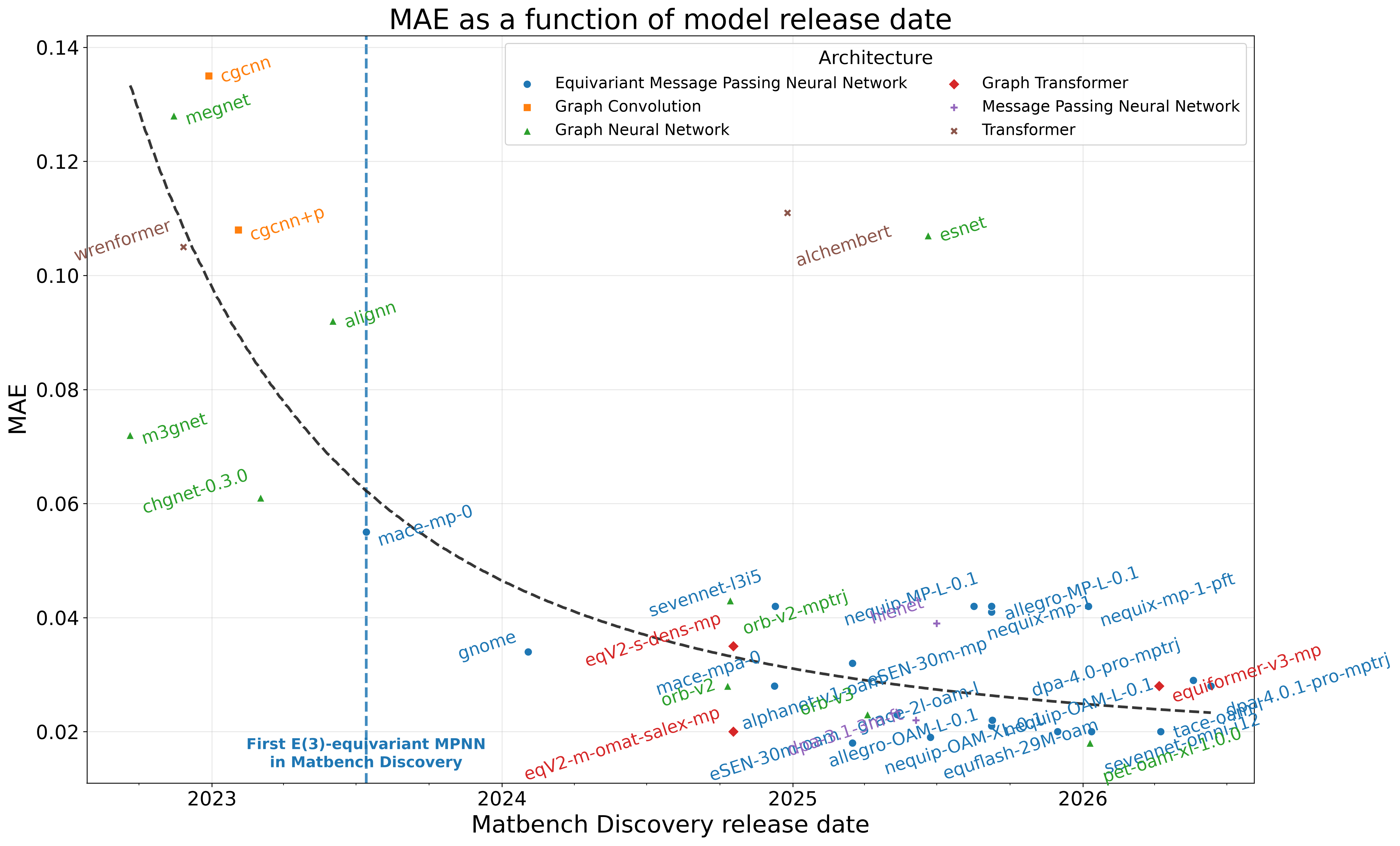}
        \caption{Model performance as a function of the Matbench Discovery
        release date. Colors indicate the architecture type.}
        \label{fig:performance-release-date}
    \end{subfigure}

    \caption{Relationship between model performance, model size, and release
    date. MAE here refers to the mean absolute error on the convex hull distance regression task. Lower MAE values indicate better performance.}
    \label{fig:performance-comparison}
\end{figure*}

Having such a knowledge graph enables us to query and explore the landscape of MLIP models in a more structured way. For example, SPARQL queries can be used to identify the most frequently used training datasets, compare model architectures, and see the relationship between model size and its predictive performance. This is what is shown, for example, in Figure~\ref{fig:performance-comparison}(a), where one of the performance metrics used in Matbench Discovery is plotted against the size of models in terms of the number of parameters. Unsurprisingly, we can see a general trend by which larger models tend to obtain better performance, with pet-oam-xl-1.0.0 being the largest and obtaining the lowest MAE, and cgcnn being among the smallest and obtaining the highest MAE on this task. Clear outliers with respect to this trend can, however, be observed, such as alchembert being a relatively large model, obtaining relatively poor performance. 

Figure~\ref{fig:performance-comparison}(b) is similar to Figure~\ref{fig:performance-comparison}(a), plotting performance according to the same criteria but against the date at which a model was added to the Matbench Discovery leaderboard. Here again, the downward trend (with lower and lower differences) is not surprising, as it indicates that newer models tend to perform better than older ones. There is also a clear correlation between the year of release of a model and its size (models are becoming bigger with time). While we can again notice significant outliers to the trend here in time, it is, however, more pronounced than the one above, showing that recency tends to be a better indicator of performance than size. \\

%As shown in Figure~\ref{fig:performance-comparison}, the MAE generally decreases as a tendency as the number of trainable parameters increases. The largest improvement is observed between the smallest models and models containing approximately $10^7$ parameters. Beyond this scale, most models achieve MAE values between approximately 0.02 and 0.04, suggesting diminishing returns from further increases in model size. 

%% perf vs architecture
\begin{figure}[ht]
    \centering
    \includegraphics[
        width=1.0\linewidth
    ]{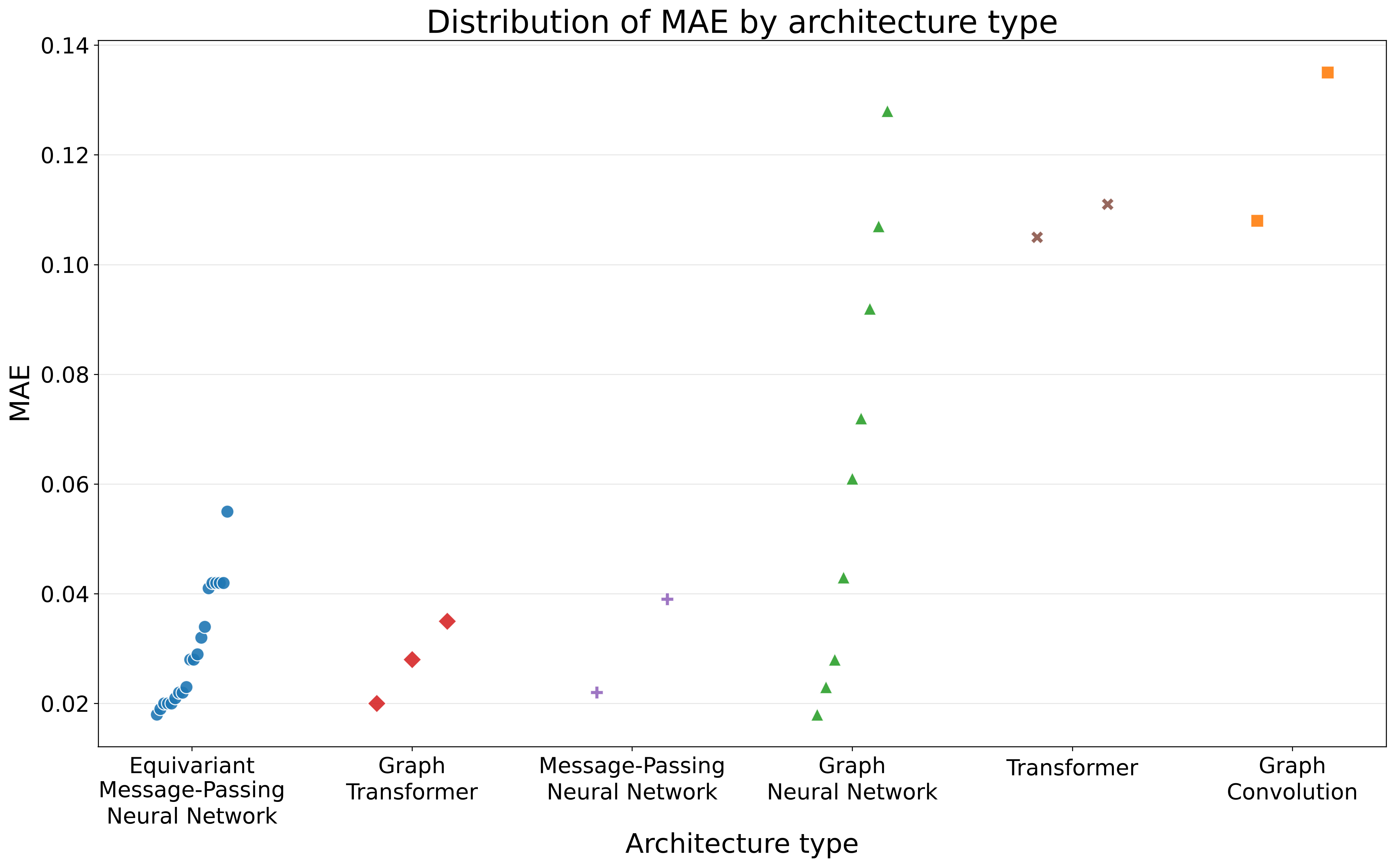}
    \caption{Model performance (MAE) across architecture types on the convex hull distance regression task. Lower values indicate better predictive performance.}
    \label{fig:performance_architecture}
\end{figure}

Both previous subfigures of Figure~\ref{fig:performance-comparison} also show the architecture type used by each model (through colors). This provides interesting information; for example, as marked on Figure~\ref{fig:performance-comparison}(b), regarding the time at which specific types of architecture appeared (here E(3)-equivariant MPNN) and how they evolved over time.  We can also see, for example, how early attempts at graph convolution were not pursued, while, in contrast, general graph neural network approaches remained in use and continued to improve (with exceptions). We can also notice the late appearance of graph transformer approaches, with models placing themselves among the best of their time. An interesting anecdotal point is that it appears from both figures that the two attempts at using transformer architectures did not seem to be particularly successful on this specific task, especially considering their size. While we provide these visualizations and discussions to illustrate the value of the knowledge graph, it shows that it could be interesting to further explore these aspects, looking at other prediction tasks and specific sets of materials. 

The relationship between model architecture and performance is depicted more precisely in Figure~\ref{fig:performance_architecture}, showing the distribution of MAE for every architecture type. This somehow confirms what was already seen previously, including that the few cases of applications of graph convolution, while effective at the time, were not considered sufficiently promising and were not pursued further. Similarly, pure transformer based architectures did not, at this point, achieve competitive performances. On the other hand, more recent innovations such as equivariant message passing architectures and graph transformers perform consistently well on the considered task, even though it can be noted that more general architectures such as message passing neural networks and general GNNs can also achieve good performance. In our process, the GNN architecture was attributed to any model that relied on a graph neural network that could not be attributed to a more specific one (such as equivariant message passing neural networks or graph transformers). This explains why, in terms of performance, this architecture ranges from among the worst to among the best, as it can include models that are built in significantly different ways. Having said that, all the graph-based methods, except for graph convolution, have models that have achieved competitive performances, with MAEs at or below 0.02, showing that other aspects (such as the size or the training dataset used) might be more important than the architecture. \\

%% most used datasets for training
\begin{figure*}[ht]
    \centering

    \begin{minipage}[t]{0.38\textwidth}
        \centering
        \vspace{0pt}
        \input{figures/dataset}
    \end{minipage}
    % \hfill
    \begin{minipage}[t]{0.58\textwidth}
        \centering
        \vspace{0pt}
        \includegraphics[
            width=\linewidth
        ]{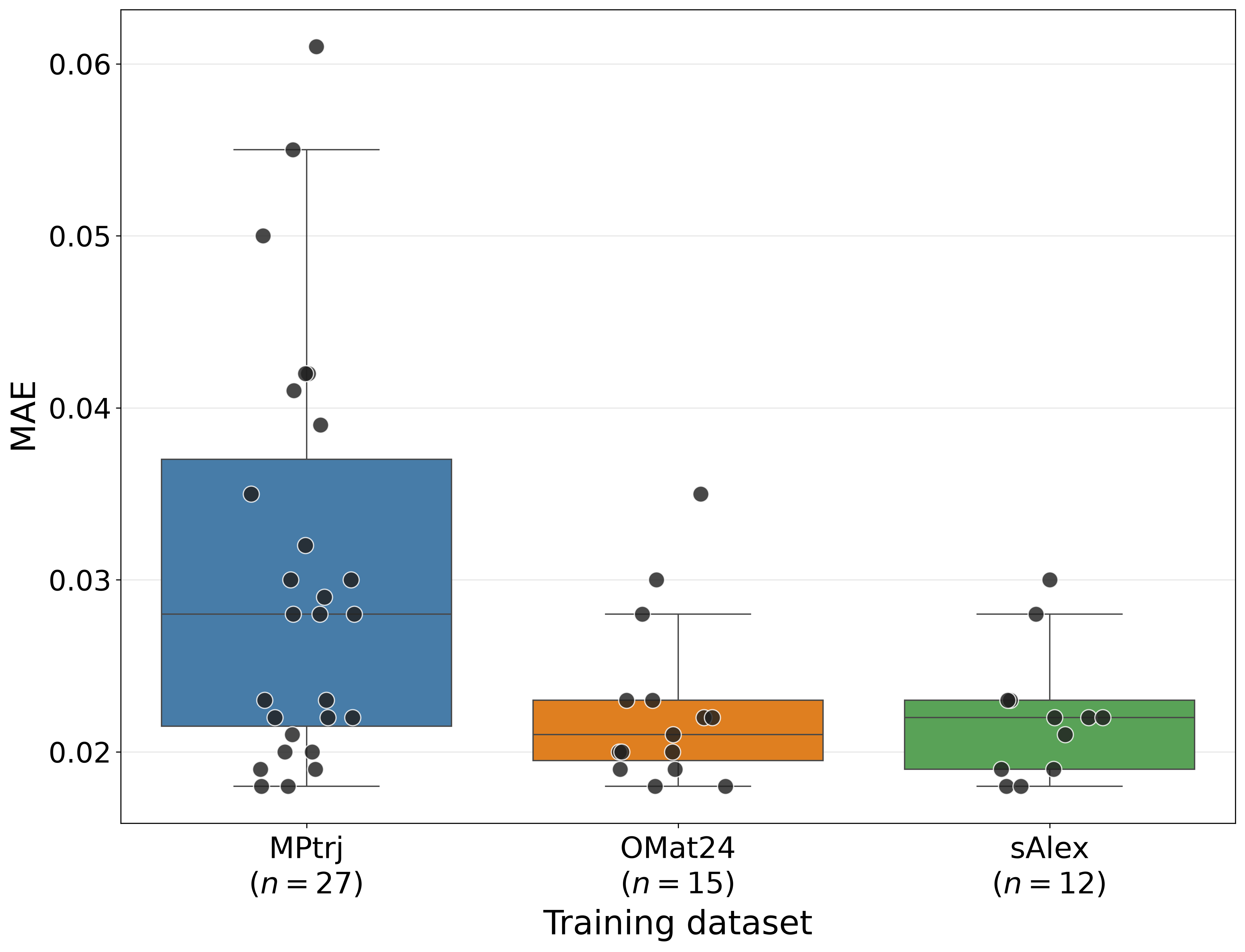}
    \end{minipage}

    \caption{Training-dataset usage and associated predictive performance. Left: the five most frequently used training datasets, measured by the number of distinct model variants. Right: distribution of MAE for models trained on MPtrj, OMat24, and sAlex. Models trained on multiple datasets are included in each corresponding group.}
    \label{fig:dataset_analysis}
\end{figure*}

Regarding training datasets, Figure~\ref{fig:dataset_analysis} shows information about the most commonly used datasets for pre-training, fine-tuning, or other training processes in the models considered here. It should be noted that a number of models use more than one dataset for training, which explains why the sum of the counts of dataset usage is higher than the number of models included in the knowledge graph. The dataset that is, by far, the most commonly used is MPtrj, which consists of extracts of information, structures, and properties from the Materials Project\footnote{\url{https://next-gen.materialsproject.org/}}. OMat24\footnote{\url{https://huggingface.co/datasets/facebook/OMAT24}} and sAlex\footnote{a subset of Alexandria - \url{https://alexandria.icams.rub.de/}} come second and third, but it could be easily noticed from the Matbench Discovery leaderboard that those three datasets are often used together. This is actually visible in the figure showing the distribution of performances (again, in terms of MAE, right of Figure~\ref{fig:dataset_analysis}) according to the dataset used. From this, it can be seen that, on average, models trained with OMat24 and sAlex achieve significantly better performances than those trained on MPtrj alone. However, it can also be seen that many of the models trained on sAlex were also trained on OMat24 and on MPtrj, while a large number of models were trained on MPtrj alone. Rather than suggesting that OMat24 and sAlex are better datasets, this seems to indicate that training on multiple datasets tends to be a more successful approach than focusing only on one. 

\section{Conclusion}

In this paper, we have described a process by which we can automatically and incrementally build a knowledge graph of machine learning models for material property predictions, focusing especially on MLIP models as present in the Matbench Discovery leaderboard. This approach includes several innovative aspects, from the recording of evidence for specific elements of information (triples in the knowledge graphs) corresponding to parts of documents in which the information was found by an LLM, to an LLM and SHACL-based validation/correction loop that ensures that the structure through which models, their training, and their evaluation are described in the knowledge graph is consistent across models. We illustrated this process by creating a knowledge graph of the models present in the Matbench Discovery leaderboard and querying it to build interesting visualizations of various aspects of those models and their performance.

There are many aspects of the proposed process that could still be improved, especially regarding its validation. Indeed, the SHACL-based loop put in place, while effective in guaranteeing an appropriate and consistent structure for the information extracted by LLMs, cannot guaranty that the information in question is factually correct. Including specific evidence for extracted triples in the knowledge graph through RDF-star can be helpful for manual validation, but this remains a complicated and time consuming task. Gold standard-based evaluations here are not feasible since building the part of the knowledge graph describing a model is an excessively complex and time consuming task. Part of our plan is therefore to make such a knowledge graph public, together with tools to support building visualizations, analyzes, and recommendations that will be attached to community feedback mechanisms. Another part is to extend the framework to include LLM-based validation and LLM ensembles to minimize the hallucination rate and the risks of incorrect content. Such methods would also enable us to identify low confidence statements on which to focus (community-based) human validation. 

Finally, another avenue for future work actually relates to those tools and, more broadly, the use of such a knowledge graph. Indeed, we illustrated some of the ways in which the knowledge graph could be queried to analyze how some factors might affect the performance of models on a specific task (convex hull distance regression). There are many tasks that could be considered, along with many other factors that could be analyzed for such investigations, which we hope a future public release of the knowledge graph will enable. One aspect that appears particularly interesting is to consider performance in relation to the characteristics of the materials being studied. This is what motivated our inclusion of the specific predictions of each model on materials included in test datasets. Indeed, metrics such as MAE are aggregated over all tested materials, but it can be expected that (in particular, depending on their training set), some models would perform differently on different subsets of materials. Assessing the relationship between materials' properties and the ability of models to accurately predict other properties for those materials seems valuable not only to better understand the models but also to identify the most suitable ones for specific tasks on specific sets of materials. A key next step for the work presented here is, therefore, to build such a recommendation engine for material prediction models. 

\section*{Declaration of use of Generative AI}

Generative AI was used within the process described in this paper (LLM based information extraction and correction), as well as in support of the development activities. Generative AI models were also used for grammar and stylistic corrections/improvements of the text of this paper.

\bibliography{semats2026}

\appendix

\end{document}

%% file: figures/dataset.tex
\begin{tikzpicture}

\begin{axis}[
    width=0.75\linewidth,
    height=1.2\linewidth,
    xlabel={Number of models},
    ylabel={Training dataset},
    label style={
        font=\large
    },
    ylabel shift=-3pt,
    xbar,
    bar width=18pt,
    xmin=0,
    xmax=32,
    symbolic y coords={
        SPICE 2,
        MP,
        sAlex,
        OMat24,
        MPtrj
    },
    ytick=data,
    yticklabel style={
        font=\small
    },
    xticklabel style={
        font=\large
    },
    grid=major,
    xmajorgrids=true,
    ymajorgrids=false,
    grid style={gray!20},
    enlarge y limits=0.14,
    nodes near coords,
    nodes near coords align={horizontal},
    every node near coord/.append style={
        font=\normalsize,
        anchor=west
    }
]

\addplot[
    fill=blue!55,
    draw=blue!70!black
]
table[
    col sep=comma,
    x=numberOfModels,
    y=datasetName
]{figures/ds.csv};

\end{axis}

\end{tikzpicture}